\pdfoutput=1
\documentclass{article}
\usepackage{natbib}

\usepackage[final]{acl}
\usepackage{times} 

\usepackage{amsmath,amsfonts,bm}

\def\eqref#1{equation~\ref{#1}}

\def\1{\bm{1}}

\DeclareMathAlphabet{\mathsfit}{\encodingdefault}{\sfdefault}{m}{sl}
\SetMathAlphabet{\mathsfit}{bold}{\encodingdefault}{\sfdefault}{bx}{n}

\usepackage[utf8]{inputenc} %
\usepackage[T1]{fontenc}    %
\usepackage{hyperref}       %
\usepackage{url}            %
\usepackage{booktabs}       %
\usepackage{amsfonts}       %
\usepackage{nicefrac}       %
\usepackage{microtype}      %
\usepackage{xcolor}         %
\usepackage{multirow}   
\usepackage{graphicx}
\usepackage[capitalise]{cleveref}
\usepackage{graphicx}
\usepackage{subcaption}
\usepackage{tabularx}
\usepackage{adjustbox}
\usepackage{amsmath}
\usepackage{xpatch}
\usepackage{esint}
\usepackage{algorithm}
\usepackage[noend]{algorithmic}
\usepackage{multirow}
\usepackage[normalem]{ulem}
\useunder{\uline}{\ul}{}
\newcommand{\RN}[1]{%
	\textup{\lowercase\expandafter{\it \romannumeral#1}}%
}

\title{Aligning CodeLLMs with Direct Preference Optimization} 

\author{
    Yibo Miao$^{1}$,
    Bofei Gao$^{2}$,
    Shanghaoran Quan$^{3}$, 
    Junyang Lin$^{3}$, Daoguang Zan$^{4}$, \\
    \textbf{Jiaheng Liu$^{3}$, Jian Yang$^{3}$,
    Tianyu Liu$^{3}$\thanks{Project lead.}, Zhijie Deng$^{1}$\thanks{Corresponding author.}} \\
        \textsuperscript{1}Shanghai Jiao Tong University \quad \textsuperscript{2}Peking University \\
        \textsuperscript{3}Alibaba Group \quad \textsuperscript{4}Institute of Software, Chinese Academy of Sciences \\
        \{miaoyibo,\;zhijied\}@sjtu.edu.cn,     tianyu0421@alibaba-inc.com\\
        }

\begin{document}
\maketitle
\begin{abstract}
The last year has witnessed the rapid progress of large language models (LLMs) across diverse domains. Among them, CodeLLMs have garnered particular attention because they can not only assist in completing various programming tasks but also represent the decision-making and logical reasoning capabilities of LLMs. However, current CodeLLMs mainly focus on pre-training and supervised fine-tuning scenarios, leaving the alignment stage, which is important for post-training LLMs, under-explored. This work first identifies that the commonly used PPO algorithm may be suboptimal for the alignment of CodeLLM because the involved reward rules are routinely coarse-grained and potentially flawed. We then advocate addressing this using the DPO algorithm. Based on only preference data pairs, DPO can render the model rank data automatically, giving rise to a fine-grained rewarding pattern more robust than human intervention. We also contribute a pipeline for collecting preference pairs for DPO on CodeLLMs. Studies show that our method significantly improves the performance of existing CodeLLMs on benchmarks such as MBPP and HumanEval.
\end{abstract}

\section{Introduction}
The past few years have witnessed the rapid development of large language models (LLMs)~\cite{touvron2023llama,chowdhery2023palm,achiam2023gpt}. 
LLMs have quickly been used in specific domains like medicine~\cite{thirunavukarasu2023large}, laws~\cite{sun2023short}, finance~\cite{yang2023fingpt}, etc. 
LLMs designed for solving coding tasks, referred to as CodeLLMs, are particularly noteworthy due to their potential to automate and streamline programming, including bug detection and code generation, thereby enhancing productivity~\cite{li2023starcoder,wei2023magicoder,guo2024deepseek}. 

Current research on CodeLLMs primarily focuses on the accumulation of extensive code-related corpora for pre-training, as well as the collection of diverse instruction-following datasets for supervised fine-tuning~\cite{roziere2023code,li2023starcoder,hui2024qwen2}.
Given that the alignment plays an important role in improving the reasoning ability of LLMs~\cite{openai_o1}, 
seminal works~\cite{le2022coderl,liu2023rltf,dou2024stepcoder} have proposed to enhance CodeLLMs through Proximal Policy Optimization (PPO)~\cite{schulman2017proximal}. 
However, we argue that they suffer from limitations in the definition of reward functions. 
For example, they commonly assign a fixed reward of -1 to any code snippet containing syntax errors that cannot be compiled, irrespective of the error count. As a result, code snippets with a single syntax error are treated no differently from those riddled with multiple errors. This makes the reward model fail to capture the nuanced preference distinctions among various code snippets and hence cannot efficiently guide the alignment of CodeLLMs.

This work proposes to align the CodeLLMs with the help of Direct Preference Optimization (DPO)~\cite{rafailov2023direct}. 
Unlike PPO, DPO does not explicitly define a reward model to capture preference, but alternatively uses model likelihood to represent that. 
By learning by comparing data pairs, DPO can automatically acquire the fine-grained differentiation between samples from coarse rewarding signals. 
Ideally, after training, a code snippet with few errors can be assigned a higher reward than that containing more errors. 
Compared to defining fine-grained, hand-crafted reward rules for better PPO, our DPO approach enjoys higher flexibility and can reduce the risk of reward hacking associated with flawed hand-crafted reward rules. 

\begin{figure}[t]
    \centering
\includegraphics[width=0.49\textwidth]{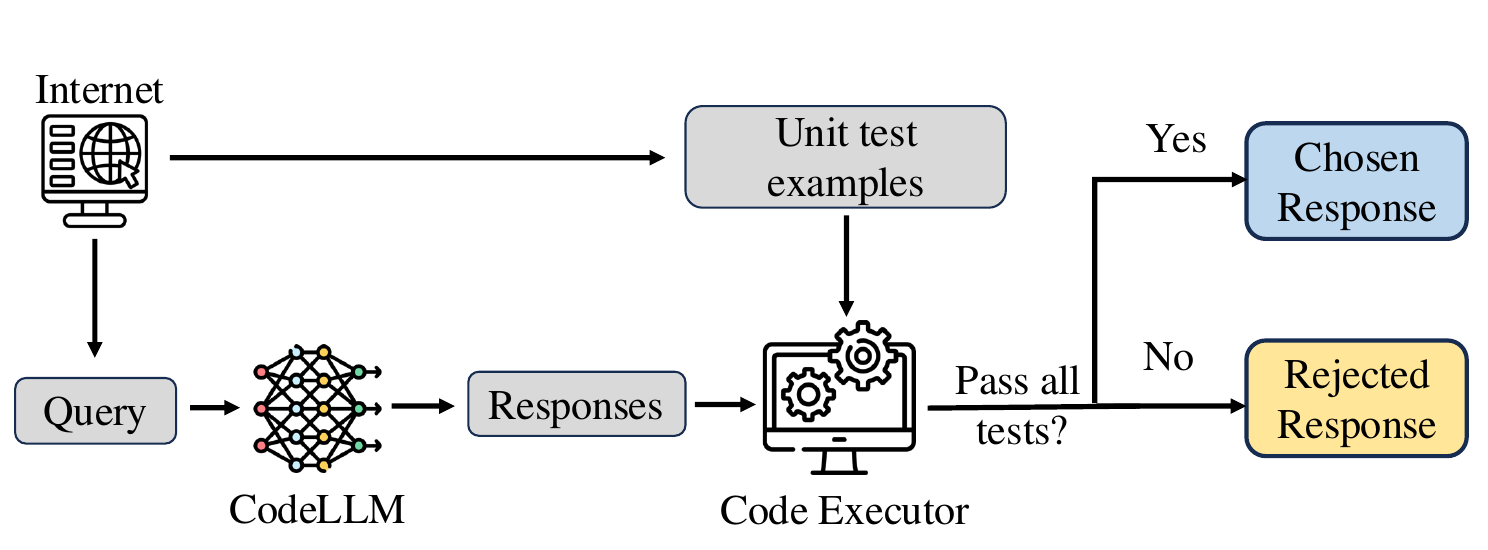}
    \caption{The pipeline of using execution feedback from the code executor to construct preference dataset.}
    \label{fig:ut_pipeline}
\end{figure}

Given that using data pairs from other models for DPO can be sub-optimal and even lead to model degradation~\cite{yan20243d,lai2024stepdpo}, 
we propose to construct on-policy preference data for DPO training, which distinguishes us from \citet{weyssow2024codeultrafeedback}. 
Specifically, we introduce external code executors to provide feedback for ranking code generations. 
An overview of this is depicted in \cref{fig:ut_pipeline}. 
Empirically, our method has demonstrably increased the performance of CodeQwen1.5 7B on MBPP~\cite{austin2021program} and HumanEval~\cite{chen2021evaluating}, enhancing the scores from 0.783 to 0.804 and 0.829 to 0.878, respectively.

\begin{figure*}[t]
    \centering
    \includegraphics[width=\textwidth]{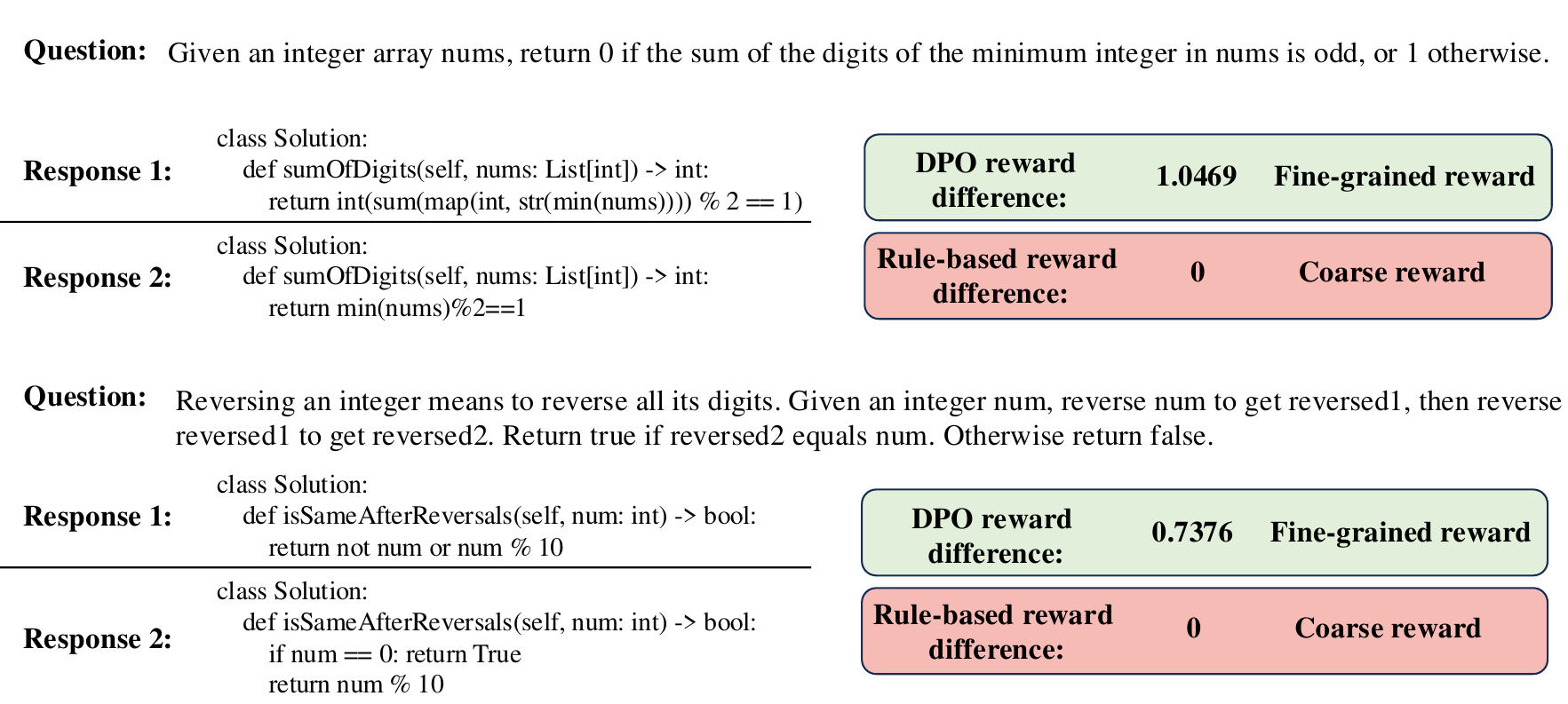}
    \caption{Two cases for illustration of the reward difference of different responses given by the DPO reward and rule-based reward. 
    When given a code question, response 1 and response 2 are two responses that have logic errors but the two responses are not the same. 
    Reward difference means the reward of response 1 minus that of response 2. 
    The rule-based reward assigns the same reward to different responses while DPO recognizes the reward difference between the different responses.}
    \label{fig:reward_margin}
\end{figure*}

\section{Methodology}
\subsection{Issues of PPO for CodeLLMs}
Let $\pi_\theta(y|x)$ denote a large language model (LLM), which generates response $y$ given the user instruction $x$.
Demonstrated by~\citet{achiam2023gpt}, 
PPO~\cite{schulman2017proximal} algorithm has been the most powerful algorithm to align LLMs for performance enhancement. 
Concretely, PPO maximizes the following objective:
\begin{equation}
  \label{ppo}
  \begin{split}
    \max_\theta J_r(\theta) = & \max_\theta \sum_{x} \mathbb{E}_{y \sim \pi_\theta(\cdot|x)}\\
    &\left[r(x, y) - \beta \log \frac{\pi_\theta(y|x)}{\pi_{0}(y|x)}\right],
  \end{split}
\end{equation}
where $r$ is the reward function,
and $\beta$ is a hyperparameter to control the degree of deviation between the policy model $\pi_\theta$ and the reference model $\pi_0$.

The effectiveness of PPO is closely linked to the quality of the reward function $r$. 
For coding tasks, we can naturally utilize the execution feedback provided by an external code executor to characterize $r$, based on specific rules.
For example, \citet{le2022coderl} define the following rewarding rule:
\begin{equation}
\label{eq:ppo_rule_reward}
r\left(x, y\right)= \begin{cases}+1, & \text { if } y \text { passed all unit tests } \\ -0.3, & \text { if } y \text { failed any unit test } \\ -0.6, & \text { if } y \text { happened runtime error } \\ -1, & \text { if } y \text { happened compile error.}\end{cases}
\end{equation}
Namely, the reward score is assigned based on the actual excitation state of the model response $y$.

Despite being widely adopted~\cite{liu2023rltf,shojaee2023execution,dou2024stepcoder}, the above rule can be too coarse and the reward space is very sparse. However, the exploration of
complex tasks like coding in an environment characterized by sparse reward is particularly challenging.
For instance, one response may contain plenty of syntax errors, while another may contain only one, yet both the two responses would receive identical rewards. 
This may cause considerable confusion for the model to align. 
Ideally, we expect the reward model to assign differentiated scores that accurately reflect the varying levels of quality among the responses.

\begin{table*}[t]
    \setlength{\tabcolsep}{12pt}
  \centering
  \begin{tabular}{lcccc}
    \hline
    \textbf{Model} & \textbf{MBPP} & \textbf{MBPP+} & \textbf{HumanEval} & \textbf{HumanEval+}\\
    \hline
    CodeQwen1.5 7B  & 0.783&	0.667&	0.829&	0.774\\
         + RFT  &0.767&	0.667&	0.793&	0.738\\
         + DPO  & \textbf{0.804}&\textbf{0.688}&	\textbf{0.878}&	\textbf{0.829}\\
     \hline
    DeepSeek-Coder 6.7B &0.743 &0.651 & 0.768&0.707\\
      + RFT &0.717&0.635&	0.768&	0.701\\
      + DPO &\textbf{0.765}&	\textbf{0.677}&	\textbf{0.787}&	\textbf{0.732}\\
      \hline
    CodeLlama-Instruct 7B & 0.548 & 0.458  &0.390 & 0.323  \\
       + RFT &0.556&0.455 & 0.360& 0.305  \\
       + DPO  &\textbf{0.571}&\textbf{0.466}&\textbf{0.427}&\textbf{0.348} \\
    \hline
  \end{tabular}
  \caption{The pass@1 perfomance of the three CodeLLMs on MBPP, MBPP+, HumanEval and HumanEval+ benchmarks. +RFT/+DPO means the CodeLLM is trained using the RFT/DPO algorithm. The \textbf{bold} indicates the optimal value.}
  \label{tab:main_result}
\end{table*}

\begin{table*}[t]
 \setlength{\tabcolsep}{12pt}
  \centering
  \begin{tabular}{lcccc}
    \hline
    \textbf{Model} & \textbf{MBPP} & \textbf{MBPP+} & \textbf{HumanEval} & \textbf{HumanEval+}\\
    \hline
    CodeQwen1.5 7B  &0.783&	0.667&	0.829&	0.774\\
          on-policy DPO &\textbf{0.804} & \textbf{0.688}&\textbf{0.878}&\textbf{0.829}\\
          off-policy DPO  & 0.799 & 0.683 & 0.860 & 0.799 \\
     \hline
      DeepSeek-Coder 6.7B &0.743 &0.651 & 0.768&0.707\\
          on-policy DPO &\textbf{0.765} & \textbf{0.677}&\textbf{0.787}&\textbf{0.732}\\
          off-policy DPO  & 0.735 & 0.640 & 0.774 & 0.695 \\
    \hline
  \end{tabular}
  \caption{The pass@1 performance of CodeLLMs after on-policy and off-policy DPO training on several benchmarks. "on-policy DPO" means the data used for training is generated by the policy model itself.  "off-policy DPO" means the data used for training is generated by another model. The \textbf{bold} indicates the optimal value.}
  \label{tab:abalation_policy}
\end{table*}

\subsection{DPO for CodeLLMs}
Instead of exploring designing more fine-grained rewarding rules in a costly trial-and-error manner, this paper proposes to utilize DPO~\cite{rafailov2023direct} to align CodeLLMs more efficiently and reliably. 
Technically, DPO operates on pre-collected preference data pairs and solves the following problem:
\begin{equation}
\label{equ:DPO loss}
\begin{split}
\mathcal{L}  = &
 - \sum_{(x, y^+, y^-)} \log \sigma \\
 &\left[\beta \log \frac{\pi_\theta({y}^+|x)}{\pi_0({y}^+|x)}
 - \beta \log \frac{\pi_\theta({y}^-|x)}{\pi_0({y}^-|x)}\right],
\end{split}
\end{equation}
where $\sigma$ denotes the sigmoid function, and $y^+$ and $y^-$ denote the preferred (i.e., chosen) response and the less preferred (i.e., rejected) one corresponding to the input $x$ respectively.
The term within the sigmoid function (i.e., subtraction between log-ratios) corresponds to an implicit reward difference learned by the model~\cite{rafailov2023direct}.

We can observe that the learning of DPO does not hinge on the exact values of the reward function $r$ but only needs the preference data pairs, which can be easily constructed given a coarse rewarding rule. 
Besides, thanks to the learning-to-rank pattern, DPO can form a fine-grained characterization of the preference difference after witnessing adequate, various data pairs. 
\cref{fig:reward_margin} provides an example of this.
In particular, we collected two incorrect responses for a certain coding problem and we expect the different responses can be assigned different rewards.
As shown, DPO can differentiate between the two responses while the hand-crafted rules assign the same reward to them.

\textbf{Details for preference pair construction.} 
We construct a set of (chosen, rejected) pairs for the training of DPO on CodeLLMs. 
As illustrated in \cref{fig:ut_pipeline}, we first aggregate a substantial number of queries (i.e., $x$) and unit test examples from the Internet, including competitive programming platforms like LeetCode\footnote{\url{https://leetcode.com}} and Codeforces\footnote{\url{https://codeforces.com}}.
Subsequently, each query is input into the CodeLLM to generate eight distinct responses, from which we extract code snippets. 
The code and the unit test examples are then submitted to the code executor for evaluation.
We assess whether the output of the code executor aligns with the ground truth delineated in the unit test examples. 
If the output is consistent with the ground truth, the response is classified as chosen; otherwise, it is deemed rejected.
For each query, we randomly select a single pair of chosen and rejected responses, yielding roughly 3,000 triples $(x,y^+,y^-)$ to support the training of DPO. 
\section{Experiments}

\subsection{Experimental Details}
\textbf{CodeLLMs of concern.} We selected CodeLLMs that have outstanding performance and have garnered wide attention from the research community, as the SFT model for further alignment. Specifically, we choose CodeQwen1.5-Instruct 7B~\cite{bai2023qwen}, CodeLlama-Instruct 7B~\cite{roziere2023code}, and DeepSeek-Coder-Instrcut 6.7B~\cite{guo2024deepseek} as the target models to conduct our experiments. 

\textbf{Benchmarks and evaluation.} In order to accurately assess the coding capabilities of CodeLLMs, we selected the most commonly used MBPP (Mostly Basic Programming Problems)~\cite{austin2021program} and HumanEval~\cite{chen2021evaluating} benchmarks. 
HumanEval comprises 164 Python problems, validated through test cases to evaluate the code produced by CodeLLMs in a zero-shot setting. 
Similarly, MBPP features 500 problems assessed in a few-shot setting.
Further, MBPP+ and HumanEval+ by~\citet{liu2024your} added dozens of times more unit test cases to the original MBPP and HumanEval benchmarks, which better reflects the ability of the CodeLLMs to truly understand and solve the coding questions. 
We also report the performance of CodeLLMs on these two benchmarks.
To ensure stable and fair comparisons, for each benchmark, we utilize the greedy decoding strategy, letting the CodeLLMs generate responses to the input code questions. 
We report the pass@1 performance across the benchmarks to illustrate the coding capabilities of the CodeLLMs.

\subsection{Main Results}
To validate the effect of DPO training, we compare the performance of CodeLLMs that use rejection sampling fine-tuning (RFT)~\cite{star,rft} and DPO for training, respectively.
Specifically, we use the chosen responses from the constructed preference dataset for RFT training and both the chosen and rejected responses from the constructed preference dataset for DPO training. 
All training hyperparameters can be acquired from the \cref{sec:hyper}.
As illustrated in \cref{tab:main_result}, while the RFT algorithm does enhance the model's performance, its improvements are not as significant as those realized through the DPO algorithm. 
This discrepancy arises because, during the optimization process with the DPO algorithm, the model is able to learn from rejected responses, enabling it to avoid generating undesirable patterns during inference and ultimately reduce errors in the generated code snippets.

\subsection{On-policy DPO vs. Off-policy DPO}
\label{sec:on_policy}
Recently, there have been some researchers who also applied DPO in the field of CodeLLM. 
\citet{weyssow2024codeultrafeedback} 
leverage the Magicoder Evol-Instruct dataset~\cite{wei2023magicoder} as the source of queries. For each query, they
randomly select four LLMs from fourteen LLMs to answer the coding problem and leverage ChatGPT to rate the four responses. 
Then they can get the preference dataset according to the rate from ChatGPT and the constructed dataset will be used for DPO training. 
However, using data generated by other models for DPO training is an off-policy mode. 
We want to emphasize that whether the preference data for DPO training is on-policy or not is very important. 
As analyzed by~\cite{lai2024stepdpo,yan20243d}, off-policy training may lead to the degradation of the model to be optimized.
Therefore, directly using the responses generated by other models for DPO may not be a good choice. 
The recommended choice is to construct the on-policy dataset using the policy model. 

To validate the argument, we set up our experiments as follows: We first used CodeQwen1.5 7B and DeepSeek-Coder 6.7B to construct preference datasets, respectively. 
In the DPO training stage, we exchange the training data between Deepseek-Coder and CodeQwen1.5. 
In other words, we use data generated by one model to train another model.
The performances of CodeLLMs after DPO training are shown in \cref{tab:abalation_policy}. 
As shown, using off-policy data leads to sub-optimal results. 
Notably, when we use preference data generated by CodeQwen1.5 to train the DeepSeek-Coder, the performance after training is even worse than the original instruct model, which indicates that using off-policy data for DPO training might be harmful to the model.

\section{Related Works}
\subsection{CodeLLMs} 
With the rapid development of LLMs, domain models specifically designed for the field of coding are continuously emerging, providing great convenience for humans.
Several pre-trained LLMs have demonstrated significant potential for code
generation, including Santacoder~\cite{allal2023santacoder}, CodeGPT~\cite{lu2021codexglue}, etc.
Moreover, CodeLLMs that underwent fine-tuning demonstrated more competitive performance, with standout models including Starcoder~\cite{li2023starcoder}, CodeLlama~\cite{roziere2023code}, Wizardcoder~\cite{luo2023wizardcoder}, DeepSeek-Coder~\cite{guo2024deepseek}, and Qwen-Coder~\cite{hui2024qwen2}.
However, when compared with the state-of-the-art CodeLLMs such as Claude-3.5-Sonnet~\cite{claude_2024} and GPT-4o~\cite{gpt_4o}, the capabilities of these models lag significantly behind.
A reason is that these CodeLLMs heavily rely on pre-train and SFT, either lacking alignment or not performing well. 

\subsection{Execution Feedback from Code Executor}
Reinforcement learning from human feedback has proven to be effective~\cite{achiam2023gpt}, but it is highly labor-intensive. 
Fortunately, the unique nature of coding tasks allows us to leverage execution feedback from a code executor to evaluate whether the code generated by CodeLLMs meets the problem's requirements. 
For example, some works use execution feedback from the code executor in inference time, for instance, ~\citet{zhong2024debug} leverage the feedback signal from the code executor to help locate the bug in the code snippet, and iteratively refine the generated code until it passes all test examples. 
Some works leverage the execution feedback from the code executor to provide a reward signal according to certain rules~\cite{liu2023rltf,dou2024stepcoder} and then PPO is applied to align the CodeLLMs for better performance.
In this paper, we propose an alternative approach to leveraging execution feedback: utilizing it to construct data for DPO training.

\section{Conclusion}
In this paper, we highlight that current CodeLLMs primarily focus on the pre-training and supervised fine-tuning stages, while neglecting the potential of the alignment stage. The existing works on using PPO to align CodeLLMs may suffer from the issue of coarse reward definition.
Therefore, we propose an approach to further enhance the ability of CodeLLMs by leveraging the execution feedback from the code executor to construct a preference dataset for DPO training. 
Moreover, we have also demonstrated that, for coding tasks, compared to off-policy DPO, it is more beneficial to adopt on-policy DPO.
In conclusion, this work proposes a practical method that can be directly used to improve the coding performance of models.

\section{Limitations}
One limitation of this work is that the relatively small number of coding problems available on the internet restricts us to constructing a limited set of preference data pairs for DPO training.
Due to this constraint, we were unable to investigate the impact of the size of the training data on the model's final performance during the alignment phase. 
Future work can be done by exploring synthesizing coding questions for data augmentation.

\bibliography{main}

\newpage
\appendix
\section{The Underlying Reward of DPO}
The DPO algorithm drives that if $\pi_\theta$ can maximize the \cref{ppo}, then the underlying reward can be given by:
\begin{equation}
\label{eq:dpo_reward}
    r(x, y)=\beta \log \frac{\pi_\theta(y \mid x)}{\pi_{\mathrm{0}}(y \mid x)}+C(x),
\end{equation}
where $C: \mathcal{X} \rightarrow \mathbb{R}$ is a scalar function. $\pi_{\mathrm{\theta}}$ is the policy model and $\pi_{\mathrm{0}}$ is the reference model.

\section{Hyperparameter Setting}
The hyperparameters used for RFT and DPO are shown in \cref{tab:hyper_parem_rft} and \cref{tab:hyper_parem_dpo}, respectively.
\label{sec:hyper}
\begin{table}
  \centering
  \begin{tabular}{cc}
    \hline
    \textbf{Hyperparameter} & \textbf{Value}\\
    \hline
     Training epochs & 1 \\
     Learning rate & 5e-6 \\
     Learning rate schedule & cosine \\
     warmup ratio & 0.05 \\
     batch size & 16 \\
     \hline
  \end{tabular}
  \caption{Hyperparameters of RFT.}
  \label{tab:hyper_parem_rft}
\end{table}

\begin{table}
  \centering
  \begin{tabular}{cc}
    \hline
    \textbf{Hyperparameter} & \textbf{Value}\\
    \hline
     Training epochs & 1 \\
     Learning rate & 5e-6 \\
     Learning rate schedule & cosine \\
     warmup ratio & 0.05 \\
     batch size & 16 \\
     beta & 0.1 \\
     \hline
  \end{tabular}
  \caption{Hyperparameters of DPO.}
  \label{tab:hyper_parem_dpo}
\end{table}

\end{document}